# Hallucination-Free Automatic Question & Answer Generation for Intuitive Learning

Nicholas X. Wang
*Stellar Learning Technologies*
Santa Clara, CA
nicholas@stellarlearning.app

Aggelos K. Katsaggelos
*Northwestern University*
Evanston, IL
a-katsaggelos@northwestern.edu

*Abstract*— Hallucinations in large language models (LLMs), defined as fluent yet incorrect or incoherent outputs, pose a significant challenge to the automatic generation of educational multiple-choice questions (MCQs). We identified four key hallucination types in MCQ generation: reasoning inconsistencies, insolvability, factual errors, and mathematical errors. To address this, we propose a hallucination-free multi-agent generation framework that breaks down MCQ generation into discrete, verifiable stages. Our framework utilizes both rule-based and LLM-based detection agents, as well as hallucination scoring metrics to optimize question quality. We redefined MCQ generation as an optimization task minimizing hallucination risk while maximizing validity, answerability, and cost-efficiency. We also introduce an agent-led refinement process that uses counterfactual reasoning and chain-of-thought (CoT) to iteratively improve hallucination in question generation. We evaluated a sample of AP-aligned STEM questions, where our system reduced hallucination rates by over 90% compared to baseline generation while preserving the educational value and style of questions. Our results demonstrate that structured multi-agent collaboration can mitigate hallucinations in educational content creation at scale, paving the way for more reliable LLM-powered learning tools.

## I. Introduction

As educational technology becomes increasingly integrated into classrooms and self-study environments, artificial intelligence (AI) is emerging as a powerful tool in the learning process [1-4]. One of the most promising applications of AI in education is automatic question generation (AQG) [5-9], which helps address the persistent shortage of high-quality practice questions-particularly in standardized test preparation and STEM education. With the rise of large language models (LLMs), such as OpenAI's ChatGPT, more students are turning to AI-driven platforms [10-15] to supplement their learning with on-demand questions and explanations.

Recognizing this potential, organizations like Khan Academy have introduced tools like Khanmigo [12], an AI-powered tutor designed to engage students in step-by-step reasoning to promote deeper understanding and critical thinking. Despite these advancements, one fundamental challenge remains: hallucinations [16-18], instances where the AI produces incorrect, misleading, or logically inconsistent outputs. In the context of question generation, hallucinations can take the form of mismatched answers and explanations, unsolvable problems, factual inaccuracies, or computational errors. These "silly mistakes" not only undermine the reliability of AI-generated content but also force students to spend additional time verifying answers, which decreases both trust and overall learning efficiency.

To mitigate hallucinations, various techniques have been proposed. One widely used approach is the Chain of Thought (CoT) prompting method [19], which encourages models to break down reasoning into intermediate steps. Leading models such as DeepSeek R1 and GPT o3 leverage CoT to improve coherence and accuracy in multi-step problems. However, even with CoT, hallucinations persist-especially in high-stakes academic contexts where precision is paramount. CoT suffers from two core limitations: First, it does not inherently verify correctness-a flawed line of reasoning can still lead to an incorrect conclusion, and the model lacks a mechanism to self-correct without external feedback. Second, the application of CoT at scale-especially with high-performing models like GPT-4 or DeepSeek R1-is computationally expensive, making it impractical for low-cost educational deployments where large volumes of content are generated daily. Thus, while CoT provides structure, it does not guarantee truthfulness, nor does it prevent subtle or domain-specific errors, such as misalignment between answers and explanations, invalid logic, or misuse of external knowledge.

This ongoing challenge underscores the need for a more robust and modular solution that can adaptively verify and refine outputs based on the type and severity of hallucination. In this paper, we introduce a novel framework designed to systematically detect and reduce hallucinations in AI-generated multiple-choice questions (MCQs). Our approach unifies Chain-of-Thought reasoning, optimization-based hallucination scoring, and an iterative multi-agent refinement architecture. Rather than relying solely on a monolithic model's output, our method dynamically applies structured verification logic based on the type of error detected-enabling early stopping, targeted rewriting, or escalation to a more capable agent when needed.

By formalizing hallucination types (e.g., logical inconsistency, unsolvable framing, factual inaccuracy, and mathematical error)

and embedding them into an optimization objective, we are able to quantify and reduce hallucination severity during generation. The multi-agent structure introduces redundancy and feedback, where specialized agents collaboratively assess answer-explanation alignment, check factual grounding, and simulate student-like reasoning to catch subtle flaws that often go undetected in traditional pipelines.

In this paper, our primary contribution is the development of a multi-agent hallucination reduction framework for AI-generated multiple-choice questions. Unlike traditional single-pass generation or basic Chain-of-Thought prompting, our system introduces a lightweight ensemble of specialized agents that iteratively verify and refine question-answer-explanation triples. Each agent targets a specific class of hallucination, such as logical inconsistency, factual inaccuracy, or computational error, and contributes to a shared scoring function that quantifies hallucination severity. By optimizing this score across refinement steps, the system dynamically improves content quality through targeted rewriting, early stopping, or escalation to stronger agents. This modular, cost-effective approach achieves significant reductions in hallucination rates while remaining scalable for real-world educational deployments.

## II. Problem Formulation

Hallucination remains one of the most significant challenges limiting the widespread deployment of LLMs in high-stakes or educational settings. In the context of question generation, these errors can lead to misinformation, confusion, and erosion of trust in automated educational systems. Addressing hallucination effectively would greatly enhance the reliability and scalability of AI-generated content, especially in domains like AP exam preparation, STEM education, and personalized tutoring. A robust solution to this problem could unlock the full potential of LLMs as intelligent educational assistants capable of generating high-quality, pedagogically sound content without human intervention.

Hallucination, in the context of large language models (LLMs), refers to any instance in which a generated output contains incorrect, misleading, or fabricated information. In the domain of educational question-answering and generation, hallucinations can significantly undermine the quality and reliability of generated content. Based on an analysis of existing failures in automated question generation, we identify four primary types of hallucinations:
1. Inconsistency between answer and explanation, where the justification provided contradicts the correct answer;
2. Impossible Question, where the question contains unsolvable or undefined elements;
3. Factual Error, where external information or domain-specific facts are misrepresented
4. Mathematical Error, where logical or computational steps are flawed.

To evaluate and mitigate these issues, we propose a structured metric for hallucination detection that systematically checks each of these four dimensions. Each generated multiple-choice question (MCQ) can be assessed across these categories, and a corresponding hallucination score can be assigned based on the severity and type of error identified. This scoring approach not only enables precise feedback for refinement but also facilitates automated quality control at scale.

We define multiple choice question (MCQ) as a tuple with $q$ as the question, $C$ as the 4 answer choices, $a$ as the answer, and $e$ as the explanation, as shown in equation (1).

$$MCQ = (q, C, a, e) \tag{1}$$

We can then define Hallucination as $H$, as shown in equation (2), compromised by the weighted sums, denoted by $\omega_i \cdot H_i$. Here, we define $\sum_{i=1}^{4} \omega_i = 1$. To be specific, $H_1$ represents inconsistency between the answer and the explanation. $H_2$ represents if the question contains unsolvable or undefined elements. $H_3$ represents factual errors, where external information or domain-specific facts are misrepresented. $H_4$ represents a mathematical error, where there is an issue with computation or logic.

$$H = \omega_1 H_1 + \omega_2 H_2 + \omega_3 H_3 + \omega_4 H_4 \tag{2}$$

We can denote $H_1$ as shown in equation (3). Essentially, this checks that the explanation semantically entails the answer.

$$H_1 = \begin{cases} 1, & \neg \vDash (e \Rightarrow a) \\ 0, & otherwise \end{cases} \tag{3}$$

Let $A_{valid}$ denote the subset of valid answers. We can denote $H_2$ as shown in equation (4).

$$H_2 = \begin{cases} 1, & |A_{valid}| \neq 1 \\ 0, & |A_{valid}| = 1 \end{cases} \tag{4}$$

Let $F_E$ denote the set of factual claims in explanation $e$. $F_E$ can be written as $f_1, f_2, f_3, \ldots f_k$. Let $K$ be a reference knowledge base with proven facts. Then, $H_3$ can be denoted as shown in equation (5).

$$H_3 = \begin{cases} 1, & if\ \exists\ f_i \in F_E\ s.t.f_i \notin K \\ 0, & otherwise \end{cases} \tag{5}$$

Finally, define $T_i$ as a terminal conclusion of a given reasoning step. Define $a_i$ as a given correct final answer of reasoning. Then $H_4$ can be defined as shown in equation (6).

$$H_4 = \begin{cases} 1, & if\ any\ T_i \neq a_i \\ 0, & otherwise \end{cases} \tag{6}$$

The problem is to try to minimize $H$ and optimize it, as shown in equation (7).

$$\min(H) = min(\omega_1 H_1 + \omega_2 H_2 + \omega_3 H_3 + \omega_4 H_4) \tag{7}$$

Because $H_i$ are independent from each other, thus minimizing $H$ is equivalent to minimizing each component of $H$ and the problem is now just $min(H_i)$ for $i=1, 2, 3$ and $4$.

### III. ITERATIVE DUAL-AGENT FRAMEWORK

To solve the optimization problem defined in Eq. (7), we adopt an iterative dual-agent framework inspired by Generative Adversarial Networks (GANs) [20-21], involving a Generator and a Detector. In this setup, the Generator is responsible for creating a multiple-choice question (MCQ), including its stem, answer choices, and explanation. The Detector evaluates the generated MCQ against a predefined hallucination typology, identifying the most probable hallucination types and tracing their potential sources.

If hallucinations are detected, the Detector provides targeted feedback, prompting the Generator to revise the MCQ accordingly. This process repeats iteratively until the output satisfies all four core criteria: consistency, solvability, factual accuracy, and logical correctness.

While both components contribute to hallucination control, the Detector inherently hallucinates less than the Generator. This is because the Generator relies on sequential generation, resulting in higher entropy and increased creative variability, beneficial for producing diverse and novel questions. In contrast, the Detector performs holistic evaluation with lower entropy, leading to more deterministic assessments and fewer hallucinations.

Let the state of the MCQ at the $t$th iteration be defined as:

$$MCQ^{(t)} = (q^{(t)}, C^{(t)}, a^{(t)}, e^{(t)}) \qquad (8)$$

The iterative Generator-Detector cycle operates as follows:
$$MCQ^{(t+1)} = G(MCQ^{(t)}, H^{(t)}) \qquad (9)$$

$$H^{(t-1)} = D(MCQ^{(t+1)}) \qquad (10)$$

Let $\epsilon$ represent a very small, acceptable margin of error. The cycle terminates when either the hallucination score falls below this threshold:
$$H^{(t)} < \epsilon_1 \qquad (11)$$
Or when the reduction in hallucination score between iterations becomes negligible:

$$|H^{(t)} - H^{(t-1)}| < \epsilon_2 \qquad (12)$$

As shown in Fig. 1, the core of our system is a multi-agent workflow composed of two key agents: the Generator and the Detector. The Generator is tasked with producing the initial version of a multiple-choice question (MCQ), including its stem, options, and explanation. Then it passes the task to the first Detector Agent, which evaluates the generated question against a premade metric. If hallucinations are detected, the Detector feeds targeted feedback back to the Generator, which then revises the question accordingly. Then, the task is passed on the task to the next Detector, as recommended by the previous one. This loop continues iteratively until the output is deemed hallucination-free based on all four critical dimensions: consistency, solvability, factual accuracy, and logical correctness or there is no more improvement in Hallucination. Fig. 1 shows one example of a possible path of iteration, as the previous Detector dynamically chooses which Detector to check next, or to early terminate.

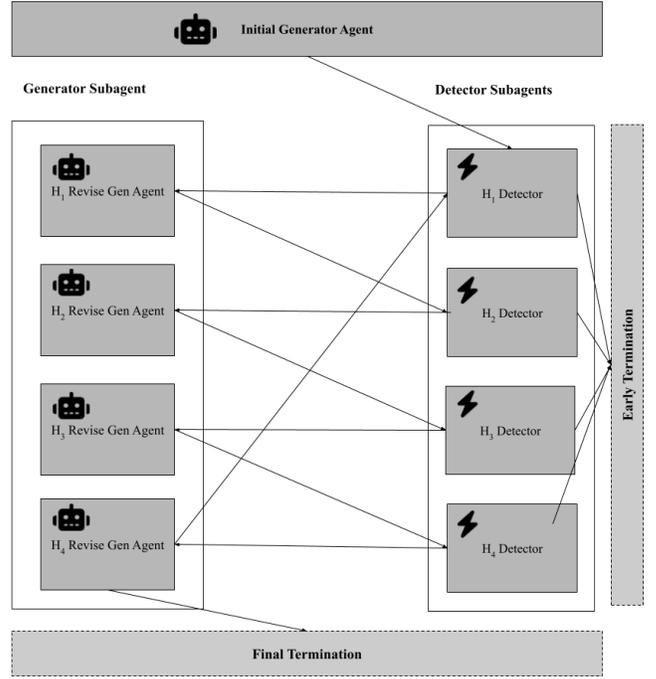

Figure 1. Iterative dual-agent framework

To enhance efficiency, our method includes dynamic iteration control, meaning that the number of refinement cycles is not fixed but adaptively determined based on hallucination severity and type. This ensures that the system maintains high accuracy while optimizing for cost. By identifying which types of hallucinations are likely to co-occur or follow from earlier mistakes, the agents can prioritize specific correction pathways, increasing the convergence rate of the system.

This multi-agent, optimized, and iterative framework represents a significant departure from conventional one-pass generation paradigms. By decomposing hallucination into its root causes and addressing them through structured agent collaboration and reasoning, our approach demonstrates that high-fidelity question generation is possible even under computational constraints. Ultimately, this opens the door for more democratized and trustworthy AI-driven educational content generation at scale.

On the other hand, our method achieves this not with top-tier models, but with more cost-efficient alternatives such as GPT-4.1-nano, thereby lowering the financial barrier to scalable and reliable question generation. While even the most advanced reasoning models, such as OpenAI's GPT-4-o, still exhibit non-negligible rates of hallucination during question generation, our framework demonstrates that hallucination can be significantly reduced, approaching zero, by applying structured refinement processes.

## IV. EXPERIMENTAL RESULTS

In our experiments, we employed GPT-4.1-nano, a lightweight and cost-effective large language model, as the foundational component of our multi-agent hallucination reduction framework. This model was selected due to its competitive performance-to-cost ratio, making it particularly suitable for scalable deployment in educational settings. As illustrated in Fig. 1, our architecture supports iterative refinement of multiple-choice questions (MCQs), with agents interacting through optimized feedback loops. Each generated question undergoes one or more verification passes depending on the detected risk of hallucination. In some cases, the process terminates early if the hallucination detector determines that additional refinement would yield negligible improvement.

Cost analysis further underscores the efficiency of our approach. GPT-4.1-nano is priced at $0.10 per 1 million tokens, in stark contrast to GPT-o3-mini, a state-of-the-art reasoning model, which costs $1.10 per 1 million tokens, a 11 times increase in token cost. Despite this disparity, our multi-agent solution not only achieves comparable or superior performance in reducing hallucinations but does so with significantly lower computational expense. Our solution can converge hallucination close to 0, by 7 iterations. Throughout the evaluation, we found that our framework consistently eliminated the majority of hallucination types, particularly those involving logical inconsistency and factual inaccuracies. These results suggest that intelligent orchestration of lightweight models, guided by structured verification strategies, can outperform more expensive reasoning-focused models on key quality metrics, all while maintaining affordability and scalability for real-world educational applications.

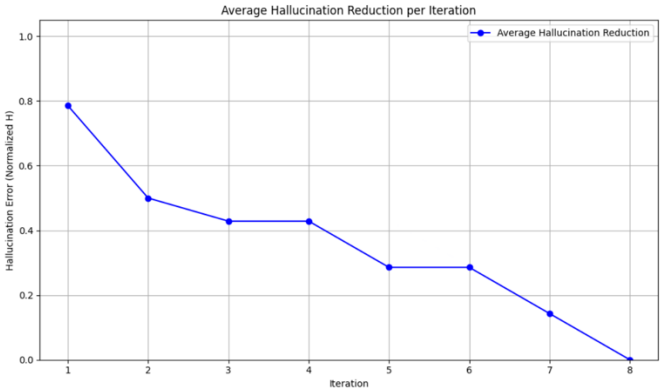

Figure 2. Hallucination Error Decreases with Iteration Increase

As illustrated in Fig. 2, hallucination rates exhibit a clear downward trend as the number of iterative refinement steps increases, ultimately converging toward nearly 0% hallucination after sufficient iterations. This demonstrates the effectiveness of our multi-agent revision process in progressively eliminating errors and inconsistencies from generated MCQs.

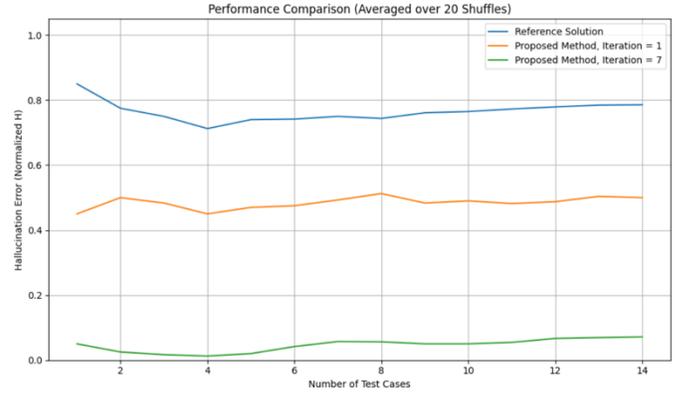

Figure 3. Hallucination Error Decreases with Iteration Increase

In Fig. 3, we observe that even a modest number of iterations yields substantial gains: hallucination decreases by approximately 50% after just one iteration compared to the reference solution (0 iterations), and by over 90% after seven iterations, indicating strong compounding benefits from repeated agent-led verification and counterfactual correction. Finally, Fig. 4 reinforces this trend by showing that as more test cases are introduced, the overall hallucination rate stabilizes at a consistently low level. This convergence supports our central hypothesis: that increased iterative scrutiny, through multi-agent refinement, reliably drives hallucination toward elimination.

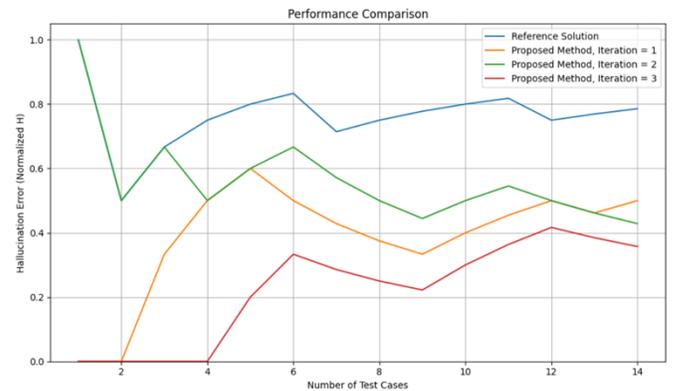

Figure 4. Hallucination Error Decreases with Iteration Increase

## V. Conclusion

Large language models show remarkable potential in generating multiple-choice questions for educational use but remain vulnerable to hallucinations that undermine trust and usability. In this work, we introduce a metric of four key hallucination types specific to the AQG domain and formalize MCQ generation as an optimization problem balancing validity, answerability, and efficiency. Our multi-agent framework reduces hallucination by decomposing generation into independent stages verified by dedicated rule-based and LLM agents. We further enhance robustness through hallucination scoring functions and counterfactual refinement loops, incorporating chain-of-thought reasoning. We conducted an experiment with a sample of AP-Aligned MCQs, which demonstrated that our approach reduces hallucination rates by over 90% while preserving authentic exam style and instructional value. These findings underscore the importance of agent-based decomposition and verification in deploying trustworthy AI-powered educational tools and open the door to future improvements via scalable scoring functions and cross-domain generalization.

## VI. Acknowledgements

We thank Neel V. Parpia, Amaar M. Chughtai, Warren Y. Li, Aaryan D. Parikh, Aditya V. Siddabathuni, and Alex H. Chen for their contributions to Stellar. We also extend our appreciation to all volunteers whose participation helped advance this research.